\documentclass[twocolumn]{bmcart}

\pdfoutput=1
\RequirePackage{natbib}
\RequirePackage{hyperref}
\usepackage[utf8]{inputenc} 
\usepackage{amsmath, amsthm, amssymb}
\usepackage{booktabs}
\usepackage{float}
\usepackage{microtype}
\usepackage{graphicx}
\usepackage{ulem}
\usepackage{color}
\usepackage{pdfpages}

\startlocaldefs
\endlocaldefs

\begin{document}
\begin{frontmatter}
  \begin{fmbox}
    \dochead{Research}

    \title{Molecular De-Novo Design through Deep Reinforcement Learning}

    \author[ addressref={aff1}, corref={aff1}, email={m.olivecrona@gmail.com} ]{\inits{MO}\fnm{Marcus} \snm{Olivecrona}} \author[ addressref={aff1}, noteref={n1},
    email={thomas.blaschke@astrazeneca.com} ]{\inits{TB}\fnm{Thomas} \snm{Blaschke}} \author[ addressref={aff1}, noteref={n1}, email={ola.engkvist@astrazeneca.com} ]{\inits{OE}\fnm{Ola}
      \snm{Engkvist}} \author[ addressref={aff1}, noteref={n1}, email={hongming.chen@astrazeneca.com} ]{\inits{HC}\fnm{Hongming} \snm{Chen}}

    \address[id=aff1]{ \orgname{Hit Discovery, Discovery Sciences, Innovative Medicines and Early Development Biotech Unit, AstraZeneca R\&D Gothenburg}, \postcode{43183} \city{Mölndal}, \cny{Sweden}
    }

\begin{artnotes}
  \note[id=n1]{thomas.blaschke@astrazeneca.com, ola.engkvist@astrazeneca.com, hongming.chen@astrazeneca.com}
\end{artnotes}

\begin{abstractbox}

\begin{abstract} 
  This work introduces a method to tune a sequence-based generative model for molecular \textit{de novo} design that through augmented episodic likelihood can learn to generate structures with certain
  specified desirable properties. We demonstrate how this model can execute a range of tasks such as generating analogues to a query structure and generating compounds predicted to be active against a
  biological target. As a proof of principle, the model is first trained to generate molecules that do not contain sulphur. As a second example, the model is trained to generate analogues to the drug
  Celecoxib, a technique that could be used for scaffold hopping or library expansion starting from a single molecule. Finally, when tuning the model towards generating compounds predicted to be
  active against the dopamine receptor type 2, the model generates structures of which more than 95\% are predicted to be active, including experimentally confirmed actives that have not been included in
  either the generative model nor the activity prediction model.

\end{abstract}


\end{abstractbox}

\end{fmbox}

\end{frontmatter}

\section{Introduction}

Drug discovery is often described using the metaphor of finding a needle in a haystack. In this case, the haystack comprises on the order of $10^{60}-10^{100}$ synthetically feasible molecules
\cite{Schneider2005}, out of which we need to find a compound which satisfies the plethora of criteria such as bioactivity, drug metabolism and pharmacokinetic (DMPK) profile, synthetic accessibility,
etc. The fraction of this space that we can synthesize and test at all - let alone efficiently - is negligible. By using algorithms to virtually design and assess molecules, \textit{de novo} design
offers ways to reduce the chemical space into something more manageable for the search of the needle.

Early \textit{de novo} design algorithms \cite{Schneider2005} used structure based approaches to grow ligands to sterically and electronically fit the binding pocket of the target of interest
\cite{Bohm1992, Gillet1994}. A limitation of these methods is that the molecules created often possess poor DMPK properties and can be synthetically intractable. In contrast, the ligand based approach
is to  generate a large virtual library of chemical structures, and search this chemical space using a scoring function that typically takes into account several
properties such as DMPK profiles, synthetic accessibility, bioactivity, and query structure similarity \cite{Reymond2013, Hartenfeller2012}. One way to create such a virtual library is to use known
chemical reactions alongside a set of available chemical building blocks, resulting in a large number of synthetically accessible structures \cite{Schneider2011}; another possibility is to use
transformational rules based on the expertise of medicinal chemists to design analogues to a query structure. For example, Besnard \textit{et al.} \cite{Besnard2012} applied a transformation rule approach to
the design of novel dopamine receptor type 2 (DRD2) receptor active compounds with specific polypharmacological profiles and appropriate DMPK profiles for a central nervous system indication. Although using either
transformation or reaction rules can reliably and effectively generate novel structures, they are limited by the inherent rigidness and scope of the predefined rules and reactions.

A third approach, known as inverse Quantitative Structure Activity Relationship (inverse QSAR), tackles the problem from a different angle: rather than first generating a virtual library and then
using a QSAR model to score and search this library, inverse QSAR aims to map a favourable region in terms of predicted activity to the corresponding molecular structures \cite{Miyao2016,
 Churchwell2004, Wong2009}. This is not a trivial problem: first the solutions of molecular descriptors corresponding to the region need to be resolved using the QSAR model, and these then need be
mapped back to the corresponding molecular structures. The fact that the molecular descriptors chosen need to be suitable both for building a forward predictive QSAR model as well as for translation
back to molecular structure is one of the major obstacles for this type of approach.

The Recurrent Neural Network (RNN) is commonly used as a generative model for data of sequential nature, and have been used successfully for tasks such as natural language processing
\cite{Mikolov2010} and music generation \cite{Eck2002}. Recently, there has been an increasing interest in using this type of generative model for the \textit{de novo} design of molecules
\cite{Segler2017, Bombarelli2016, Yu2016}. By using a data-driven method that attempts to learn the underlying probability distribution over a large set of chemical structures, the search over the
chemical space can be reduced to only molecules seen as reasonable, without introducing the rigidity of rule based approaches. Segler \textit{et al.} demonstrated that an RNN trained on the
canonicalized SMILES representation of molecules can learn both the syntax of the language as well as the distribution in chemical space \cite{Segler2017}. They also show how further training of the
model using a focused set of actives towards a biological target can produce a fine-tuned model which generates a high fraction of predicted actives.

In two recent studies, reinforcement learning (RL) \cite{Sutton1998} was used to fine tune pre-trained RNNs. Yu \textit{et al.} \cite{Yu2016} use an adversarial network to estimate the expected return for
state-action pairs sampled from the RNN, and by increasing the likelihood of highly rated pairs improves the generative network for tasks such as poem generation. Jaques \textit{et al.}
\cite{Jaques2016} use Deep Q-learning to improve a pre-trained generative RNN by introducing two ways to score the sequences generated: one is a measure of how well the sequences adhere to music
theory, and one is the likelihood of sequences according to the initial pre-trained RNN. Using this concept of prior likelihood they reduce the risk of forgetting what was initially learnt by the RNN,
compared to a reward based only on the adherence to music theory. The authors demonstrate significant improvements over both the initial RNN as well as an RL only approach. They
later extend this method to several other tasks including the generation of chemical structures, and optimize toward molecular properties such as cLogP \cite{Leo1971} and QED drug-likeness \cite{Bickerton2012}.
However, they report that the method is dependent on a reward function incorporating handwritten rules to penalize undesirable types of sequences, and even then can lead to exploitation of the reward resulting in unrealistically simple molecules that are more likely to satisfy the optimization requirements than more
complex structures \cite{Jaques2016}.

In this study we propose a policy based RL approach to tune RNNs for episodic tasks \cite{Sutton1998}, in this case the task of generating molecules with given
desirable properties. Through learning an augmented episodic likelihood which is a composite of prior likelihood \cite{Jaques2016} and a user defined scoring function, the method aims to fine-tune an
RNN pre-trained on the ChEMBL database \cite{Gaulton2012} towards generating desirable compounds. Compared to maximum likelihood estimation finetuning \cite{Segler2017}, this method can make use of
negative as well as continuous scores, and may reduce the risk of catastrophic forgetting \cite{Goodfellow2013}. The method is applied to several different tasks of molecular \textit{de novo} design,
and an investigation was carried out to illustrate how the method affects the behaviour of the generative model on a mechanistic level.

\section{Methods}
\subsection{Recurrent Neural Networks}
A recurrent neural network is an architecture of neural networks designed to make use of the symmetry across steps in sequential data while simultaneously at every step keeping track
of the most salient information of previously seen steps, which may affect the interpretation of the current one \cite{Goodfellow2016}. It does so by introducing the concept of a \textit{cell} (Figure
\ref{fig:rnn}). For any given step $t$, the $cell_{t}$ is a result of the previous $cell_{t-1}$ and the current input $x^{t-1}$. The content of $cell_t$ will determine both the output at the current
step as well as influence the next cell state. The cell thus enables the network to have a memory of past events, which can be used when deciding how to interpret new data. These properties make an
RNN particularly well suited for problems in the domain of natural language processing. In this setting, a sequence of words can be encoded into one-hot vectors the length of our vocabulary $X$. Two
additional tokens, $GO$ and $EOS$, may be added to denote the beginning and end of the sequence respectively.

\subsubsection{Learning the data}
\begin{figure}
\includegraphics{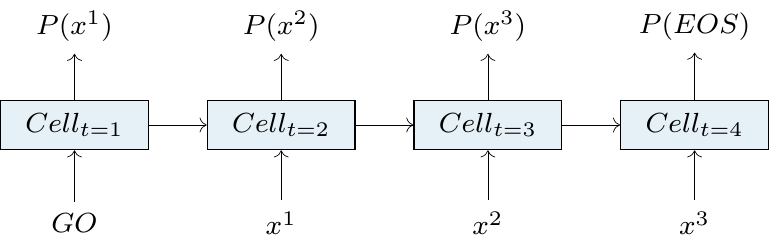}
\caption{\csentence{Learning the data.}
 Depiction of maximum likelihood training of an RNN. $x^t$ are the target sequence tokens we are trying to learn by maximizing $P(x^t)$ for each step.}
\label{fig:rnn}
\end{figure}

Training an RNN for sequence modeling is typically done by maximum likelihood estimation
of the next token $x^{t}$ in the target sequence given tokens for the previous steps (Figure \ref{fig:rnn}). At every step the model will produce a probability distribution over what the
next character is likely to be, and the aim is to maximize the likelihood assigned to the correct token:
\begin{equation*}
  J(\Theta) = -\sum_{t=1}^T {\log P{(x^{t}\mid x^{t-1},...,x^{1})}}   
\end{equation*}
The cost function $J(\Theta)$, often applied to a subset of all training examples known as a batch, is minimized with respect to the network parameters $\Theta$. Given a predicted log likelihood $\log
P$ of the target at step $t$, the gradient of the prediction with respect to $\Theta$ is used to make an update of $\Theta$. This method of fitting a neural network is called back-propagation. Due to
the architecture of the RNN, changing the network parameters will not only affect the direct output at time $t$, but also affect the flow of information from the previous cell into the current one
iteratively. This domino-like effect that the recurrence has on back-propagation gives rise to some particular problems, and back-propagation applied to RNNs is referred to as back-propagation through
time (BPTT).

BPTT is dealing with gradients that through the chain-rule contains terms which are multiplied by themselves many times, and this can lead to a phenomenon known as exploding and vanishing gradients.
If these terms are not unity, the gradients quickly become either very large or very small. In order to combat this issue, Hochreiter \textit{et al.} introduced the Long-Short-Term Memory cell
\cite{Hochreiter1997}, which through a more controlled flow of information can decide what information to keep and what to discard. The Gated Recurrent Unit is a simplified implementation of the
Long-Short-Term Memory architecture that achieves much of the same effect at a reduced computational cost \cite{Chung2014}.

\subsubsection{Generating new samples}
\label{subsec:generating samples}
\begin{figure}
\includegraphics{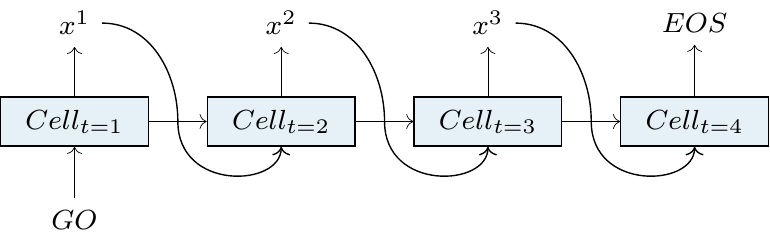}
\caption{\csentence{Generating sequences.}
 Sequence generation by a trained RNN. Every timestep $t$ we sample the next token of the sequence $x^{t}$ from the probability distribution given by the RNN, which is then fed in as the next input.}
\label{fig:rnn2}
\end{figure}

Once an RNN has been trained on target sequences, it can then be used to generate new sequences that follow the conditional probability distributions learned from the training set. The first input -
the $GO$ token - is given and at every timestep after we sample an output token $x^t$ from the predicted probability distribution $P(X^t)$ over our vocabulary $X$ and use $x^t$ as our next input. Once
the $EOS$ token is sampled, the sequence is considered finished (Figure \ref{fig:rnn2}).

\subsubsection{Tokenizing and one-hot encoding SMILES}
\begin{figure}[hb]
\includegraphics{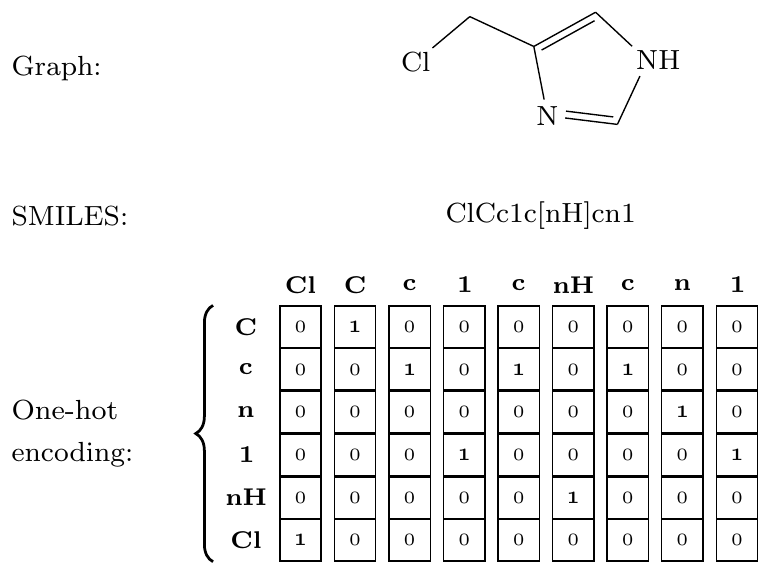}
\caption{\csentence{Three representations of 4-(chloromethyl)-1H-imidazole.} Depiction of a one-hot representation derived from the SMILES of a molecule. Here a reduced vocabulary is shown, while in practice a much larger vocabulary that covers all tokens present in the training data is used.}
\label{fig:smiles}
\end{figure}

A SMILES \cite{SMILES}  represents a molecule as a sequence of characters corresponding to atoms as well as special characters denoting opening and closure of rings and
branches. The SMILES  is, in most cases, tokenized based on a single character, except for atom types which comprise two characters such as "Cl" and "Br" and
special environments denoted by square brackets (e.g [nH]), where they are considered as one token. This method of tokenization resulted in 86 tokens present in the training data. Figure
\ref{fig:smiles} exemplifies how a chemical structure is translated to both the SMILES and one-hot encoded representations.

There are many different ways to represent a single molecule using SMILES. Algorithms that always represent a certain molecule with the same SMILES are referred to as canonicalization
  algorithms \cite{Weininger1989}. However, different implementations of the algorithms can still produce different SMILES.

\subsection{Reinforcement Learning}
Consider an Agent, that given a certain state $s\in\mathbb{S}$ has to choose which action $a\in\mathbb{A}(s)$ to take, where $\mathbb{S}$ is the set of possible states and $\mathbb{A}(s)$ is the set
of possible actions for that state. The policy $\pi(a \mid s)$ of an Agent maps a state to the probability of each action taken therein. Many problems in reinforcement learning are framed as Markov
decision processes, which means that the current state contains all information necessary to guide our choice of action, and that nothing more is gained by also knowing the history of past states. For
most real problems, this is an approximation rather than a truth; however, we can generalize this concept to that of a partially observable Markov decision process, in which the Agent can interact
with an incomplete representation of the environment. Let $r(a \mid s)$ be the reward which acts as a measurement of how good it was to take an action at a certain state, and the long-term return
$G(a_t, S_t = \sum_{t}^T{r_t}$ as the cumulative rewards starting from $t$ collected up to time $T$. Since molecular desirability in general is only sensible for a completed SMILES, we will refer only to the return of a complete
sequence.

What reinforcement learning concerns, given a set of actions taken from some states and the rewards thus received, is how to improve the Agent policy in such a way as to increase the expected return
$\mathbb{E}[G]$. A task which has a clear endpoint at step $T$ is referred to as an episodic task \cite{Sutton1998}, where $T$ corresponds to the length of the episode. Generating a SMILES is an
example of an episodic task, which ends once the $EOS$ token is sampled.

The states and actions used to train the agent can be generated both by the agent itself or by some other means. If they are generated by the agent itself the learning is referred to as
\textit{on-policy}, and if they are generated by some other means the learning is referred to as \textit{off-policy} \cite{Sutton1998}.

There are two different approaches often used in reinforcement learning to obtain a policy: value based RL, and policy based RL \cite{Sutton1998}. In value based
  RL, the goal is to learn a value function that describes the expected return from a given state. Having learnt this function, a policy can be determined in such a way as to maximize the
  expected value of the state that a certain action will lead to. In policy based RL on the other hand, the goal is to directly learn a policy. For the problem addressed in this
  study, we believe that policy based methods is the natural choice for three reasons:
  \begin{itemize}
  \item{Policy based methods can learn explicitly an optimal stochastic policy \cite{Sutton1998}, which is our goal.}
  \item{The method used starts with a prior sequence model. The goal is to finetune this model according to some specified scoring function. Since the prior model already constitutes a policy,
      learning a finetuned policy might require only small changes to the prior model.}
  \item{The episodes in this case are short and fast to sample, reducing the impact of the variance in the estimate of the gradients.}
  \end{itemize}

In Section \ref{subsec:drd2} the change in policy between the prior and the finetuned model is investigated, providing justification for the second point.

\begin{figure*}
\includegraphics{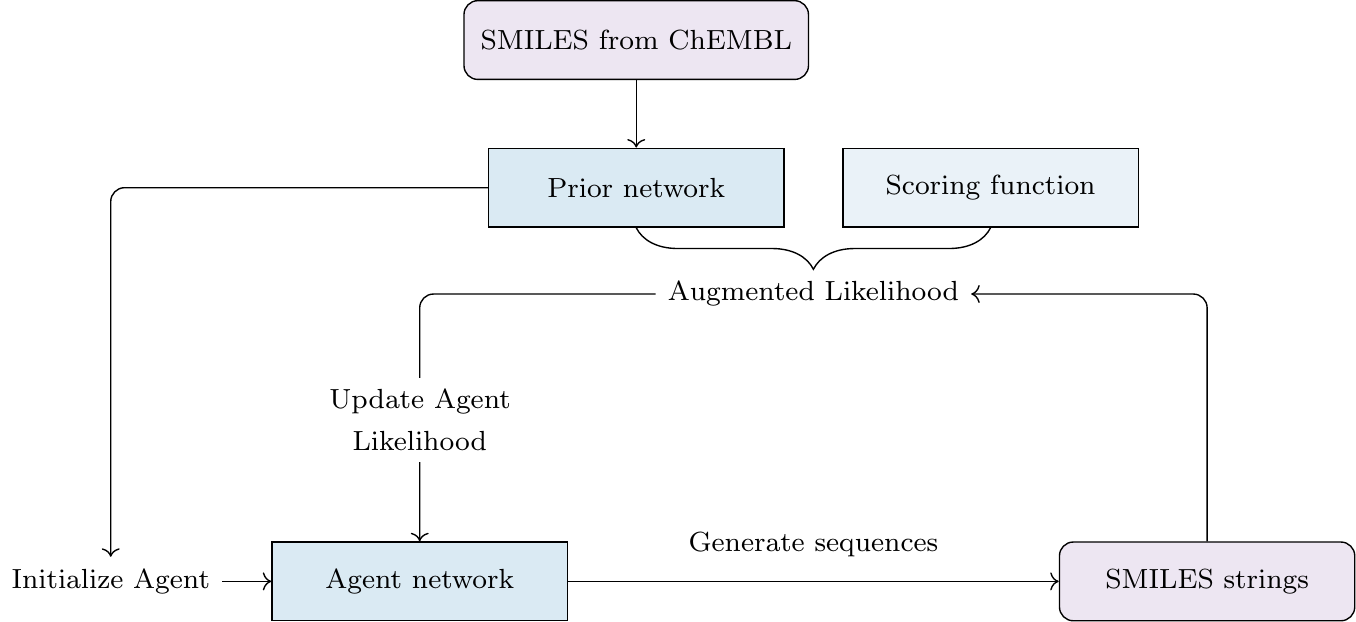}
\caption{\csentence{The Agent.}
 Illustration of how the model is constructed. Starting from a Prior network trained on ChEMBL, the Agent is trained using the augmented likelihood of the SMILES generated.}
\label{fig:agent}
\end{figure*}

\subsection{The Prior network}
Maximum likelihood estimation was employed to train the initial RNN composed of 3 layers with 1024 Gated Recurrent Units (forget bias 5) in each layer. The RNN was trained on the RDKit \cite{RDKit}
canonical SMILES of 1.5 million structures from ChEMBL \cite{Gaulton2012} where the molecules were restrained to containing between 10 and 50 heavy atoms and elements $\in\{H, B, C, N, O, F, Si, P, S,
Cl, Br, I\}$. The model was trained with stochastic gradient descent for 50 000 steps using a batch size of 128, utilizing the Adam optimizer \cite{Kingma2014} ($\beta_1 = 0.9$, $\beta_2 = 0.999$, and
$\epsilon = 10^{-8}$) with an initial learning rate of 0.001 and a 0.02 learning rate decay every 100 steps. Gradients were clipped to $[-3, 3]$. Tensorflow \cite{Tensorflow} was used to implement the
Prior as well as the RL Agent.

\subsection{The Agent network}
\label{sec:agent}
We now frame the problem of generating a SMILES representation of a molecule with specified desirable properties via an RNN as a partially observable Markov decision process, where the agent must make
a decision of what character to choose next given the current cell state. We use the probability distributions learnt by the previously described prior model as our initial prior policy. We will refer to
the network using the prior policy simply as the \textit{Prior}, and the network whose policy has since been modified as the \textit{Agent}. The Agent is thus also an RNN with the same
  architecture as the Prior. The task is episodic, starting with the first step of the RNN and ending when the $EOS$ token is sampled. The sequence of actions $A = {a_1, a_2,...,a_T}$ during this
episode represents the SMILES generated and the product of the action probabilities $P(A) = \prod_{t = 1}^T{\pi(a_t \mid s_t)}$ represents the model likelihood of the sequence formed. Let $S(A)\in[-1,
1]$ be a scoring function that rates the desirability of the sequences formed using some arbitrary method. The goal now is to update the agent policy $\pi$ from the prior policy $\pi_{Prior}$ in such
a way as to increase the expected score for the generated sequences. However, we would like our new policy to be anchored to the prior policy, which has learnt both the syntax
of SMILES and distribution of molecular structure in ChEMBL \cite{Segler2017}. We therefore denote an augmented likelihood $\log P(A)_\mathbb{U}$ as a prior likelihood modulated by the desirability of a sequence:
\begin{equation*}
  \log P(A)_\mathbb{U} = \log P(A)_{Prior} + \sigma S(A)
\end{equation*}
where $\sigma$ is a scalar coefficient. The return $G(A)$ of a sequence $A$ can in this case be seen as the agreement between the Agent likelihood $\log P(A)_\mathbb{A}$ and the augmented likelihood:
\begin{equation*}
  G(A) = -[\log P(A)_\mathbb{U} - \log P(A)_\mathbb{A}]^2   
\end{equation*}
The goal of the Agent is to learn a policy which maximizes the expected return, achieved by minimizing the cost function $L(\Theta) = -G$. The fact that we describe
  the target policy using the policy of the Prior and the scoring function enables us to formulate this cost function. In the appendix we show how this approach 
can be described using a REINFORCE \cite{Williams1992} algorithm with a final step reward of $r(t) = [\log P(A)_\mathbb{U} - \log P(A)_\mathbb{A}]^2 / \log P(A)_\mathbb{A}$. We believe this is a
more natural approach to the problem than REINFORCE algorithms directly using rewards of $S(A)$ or $\log P(A)_{Prior} + \sigma S(A)$. In Section
\ref{sec:sulphur} we compare our approach to these methods. The Agent is trained in an on-policy fashion on batches of
128 generated sequences, making an update to $\pi$ after every batch has been generated and scored. A standard gradient descent optimizer with a learning rate of 0.0005 was used and gradients were
clipped to $[-3, 3]$.

Figure \ref{fig:agent} shows an illustration of how the Agent, initially identical to the Prior, is trained using reinforcement learning. The training shifts the probability distribution from that of
the Prior towards a distribution modulated by the desirability of the structures. This method adopts a similar concept to Jaques \textit{et al.} \cite{Jaques2016}, while using a policy based
  RL method that introduces a novel cost function with
the aim of addressing the need for handwritten rules and the issues of generating structures that are too simple. 

In all the tasks investigated below, the scoring function is fixed during the training of the Agent. If instead
  the scoring function used is defined by a discriminator network whose task is to distinguish sequences generated by the Agent from `real' SMILES (e.g. a set of actives), the method could be described as a type of Generative Adversarial Network \cite{Goodfellow2014}, where the Agent and the discriminator would be jointly trained in
  a game where they both strive to beat the other. This is the approach taken by Yu \textit{et al.} \cite{Yu2016} to finetune a pretrained sequence model for poem generation.
  Guimaraes \textit{et al.} demonstrates how such a method can be combined with a fixed scoring function for molecular \textit{de novo} design \cite{Guimaraes2017}.

\begin{figure*}
\includegraphics{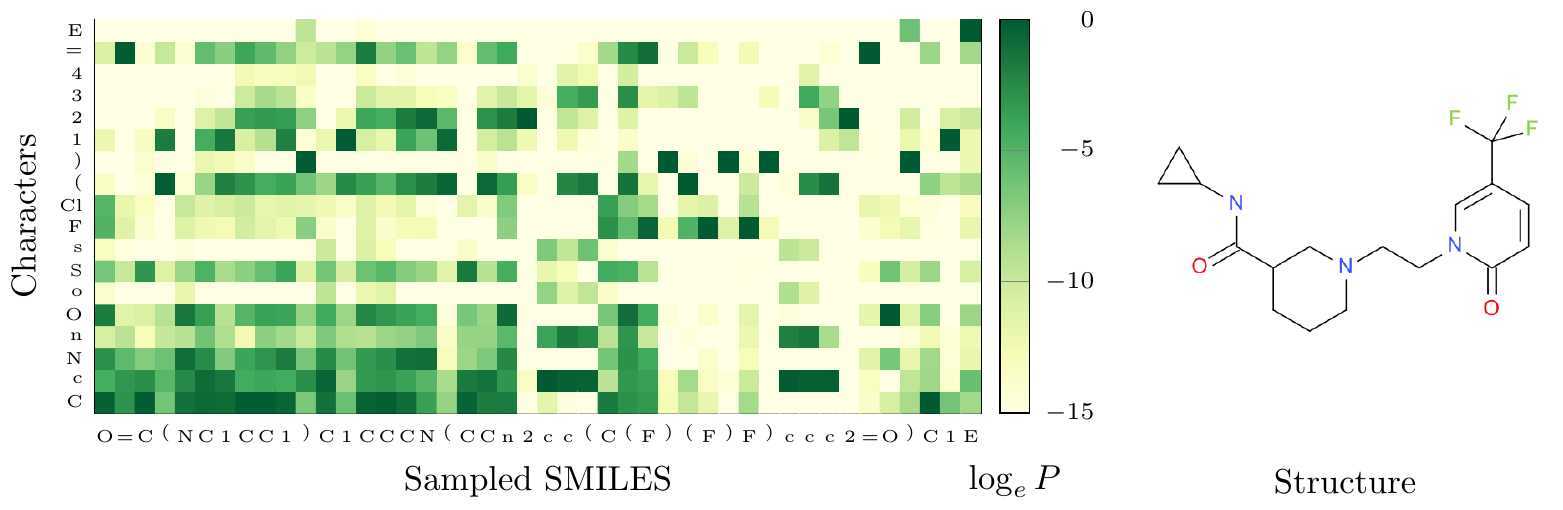}
\caption{\csentence{How the model thinks while generating the molecule on the right.} Conditional probability over the next token as a function of previously chosen ones according to the model. On the y-axis is shown the probability distribution for the character to be choosen at the current step, and on the x-axis is shown the character that in this instance was sampled. E = EOS.}
\label{fig:heatmap}
\end{figure*}

\subsection{The DRD2 activity model}
\label{sec:DRD2}
In one of our studies the objective of the Agent is to generate molecules that are predicted to be active against a biological target. The dopamine type 2 receptor DRD2 was chosen as the target, and
corresponding bioactivity data was extracted from ExCAPE-DB \cite{Sun2017}. In this dataset there are 7218 actives (pIC50 $>$ 5) and 343204 inactives (pIC50 $<$ 5). A subset of 100 000 inactive
compounds was randomly selected. In order to decrease the nearest neighbour similarity between the training and testing structures \cite{Sheridan2013, Unterthiner2014, Mayr2016}, the actives were
grouped in clusters based on their molecular similarity. The Jaccard \cite{Jaccard1901} index, for binary vectors also known as the Tanimoto similarity, based on the RDKit implementation of binary
Extended Connectivity Molecular Fingerprints with a diameter of 6 (ECFP6 \cite{Rogers2010}) was used as a similarity measure and the actives were clustered using the Butina clustering algorithm
\cite{Butina1999} in RDKit with a clustering cutoff of 0.4. In this algorithm, centroid molecules will be selected, and everything with a similarity higher than 0.4 to these centroids will be assigned
to the same cluster. The centroids are chosen such as to maximize the number of molecules that are assigned to any cluster. The clusters were sorted by size and iteratively assigned to
the test, validation, and training sets (assigned 4 clusters each iteration) to give a distribution of $\frac{1}{6}$, $\frac{1}{6}$, and $\frac{4}{6}$ of the
clusters respectively. The inactive compounds, of which less than 0.5\% were found to belong to any of the clusters formed by the actives, were split randomly into the three sets using the same
ratios.

A support vector machine (SVM) classifier with a Gaussian kernel was built in Scikit-learn \cite{scikit-learn} on the training set as a predictive model for DRD2 activity. The optimal C and Gamma
values utilized in the final model were obtained from a grid search for the highest ROC-AUC performance on the validation set.

\section{Results and Discussion}

\subsection{Structure generation by the Prior}
After the initial training, 94\% of the sequences generated by the Prior as described in Section \ref{subsec:generating samples} corresponded to valid molecular structures according to RDKit
\cite{RDKit} parsing, out of which 90\% are novel structures outside of the training set. A set of randomly chosen structures generated by the Prior, as well as by Agents trained in the subsequent
examples, are shown in the appendix. The process of generating a SMILES by the Prior is illustrated in Figure \ref{fig:heatmap}. For every token in the generated SMILES sequence, the conditional
probability distribution over the vocabulary at this step according to the Prior is displayed. The sequence of distributions are depicted in Figure \ref{fig:heatmap}. For the first step, when no
information other than the initial GO token is present, the distribution is an approximation of the distribution of first tokens for the SMILES in the ChEMBL training set. In this case "O" was
sampled, but "C", "N", and the halogens were all likely as well. Corresponding log likelihoods were -0.3 for "C", -2.7 for "N", -1.8 for "O", and -5.0 for "F" and "Cl".

A few (unsurprising) observations:
\begin{itemize}
\item{Once the aromatic "n" has been sampled, the model has come to expect a ring opening (i.e. a number), since aromatic moieties by definition are cyclic.}
\item{Once an aromatic ring has been opened, the aromatic atoms "c", "n", "o", and "s" become probable, until 5 or 6 steps later when the model thinks it is time to close the ring.}
\item{The model has learnt the RDKit canonicalized SMILES format of increasing ring numbers, and expects the first ring to be numbered "1". Ring numbers can be reused, as in the two first rings in this example.
    Only once "1" has been sampled does it expect a ring to be numbered "2" and so on.}
\end{itemize}

\subsection{Learning to avoid sulphur}
\label{sec:sulphur}

\begin{table*}[h]
  \centering
  \caption{Comparison of model performance and properties for non-sulphur containing structures generated by the two models. Properties reported as $\mathrm{Mean} \pm \mathrm{StdDev}$.}
  \begin{tabular}{lccccc}
    \toprule
    Model & Prior & Agent & Action basis & REINFORCE & REINFORCE + Prior\\ 
    \midrule
    Fraction of valid SMILES & $0.94 \pm 0.01$ & $0.95 \pm 0.01$ & $0.95 \pm 0.01$  & $0.98 \pm 0.00$ & $0.98 \pm 0.00$\\
    Fraction without sulphur & $0.66 \pm 0.01$ & $0.98 \pm 0.00$ & $0.92 \pm 0.02$ & $0.98 \pm 0.00$ & $0.92 \pm 0.01$\\
    Average molecular weight & $371 \pm 1.70$ & $367 \pm 3.30$ & $372 \pm 0.94$ & $585 \pm 27.4$ & $232 \pm 5.25$\\
    Average cLogP & $3.36 \pm 0.04$ & $3.37 \pm 0.09$ & $3.39 \pm 0.02$ & $11.3 \pm 0.85$ & $3.05 \pm 0.02$\\
    Average NumRotBonds & $5.39 \pm 0.04$ & $5.41 \pm 0.07$ & $6.08 \pm 0.04$ & $30.0 \pm 2.17$ & $2.8 \pm 0.11$\\
    Average NumAromRings & $2.26 \pm 0.02$ & $2.26 \pm 0.02$ & $2.09 \pm 0.02$ & $0.57 \pm 0.04$ & $2.11 \pm 0.02$\\
    \bottomrule
  \end{tabular}
  \label{table:sulphur}
\end{table*}

\begin{table*}[h]
  \centering
  \caption{Randomly selected SMILES generated by the different models.}
  \begin{tabular}{lccccc}
    \toprule
    Model & Sampled SMILES\\ 
    \midrule
          & CCOC(=O)C1=C(C)OC(N)=C(C\#N)C1c1ccccc1C(F)(F)F \\
    Prior & COC(=O)CC(C)=NNc1ccc(N(C)C)cc1[N+](=O)[O-] \\
          & Cc1ccccc1CNS(=O)(=O)c1ccc2c(c1)C(=O)C(=O)N2 \\
    \midrule
          & CC(C)(C)NC(=O)c1ccc(OCc2ccccc2C(F)(F)F)nc1-c1ccccc1 \\
    Agent & CC(=O)NCC1OC(=O)N2c3ccc(-c4cccnc4)cc3OCC12 \\ 
          & OCCCNCc1cccc(-c2cccc(-c3nc4ccccc4[nH]3)c2OCCOc2ncc(Cl)cc2Br)c1 \\
    \midrule
          & CCN1CC(C)(C)OC(=O)c2cc(-c3ccc(Cl)cc3)ccc21 \\
    Action level & CCC(CC)C(=O)Nc1ccc2cnn(-c3ccc(C(C)=O)cc3)c2c1 \\
          & CCCCN1C(=O)c2ccccc2NC1c1ccc(OC)cc1 \\
    \midrule
          & CC1CCCCC12NC(=O)N(CC(=O)Nc1ccccc1C(=O)O)C2=O \\
    REINFORCE & CCCCCCCCCCCCCCCCCCCCCCCCCCCCNC(=O)OCCCCCC \\
          & CCCCCCCCCCCCCCCCCCCCCC1CCC(O)C1(CCC)CCCCCCCCCCCCCCC \\
    \midrule
          & Nc1ccccc1C(=O)Oc1ccccc1 \\
    REINFORCE + Prior & O=c1cccccc1Oc1ccccc1 \\ 
          & Nc1ccc(-c2ccccc2O)cc1 \\
    \bottomrule
  \end{tabular}
  \label{table:sulphursmiles}
\end{table*}

As a proof of principle the Agent was first trained to generate molecules which do not contain sulphur. The method described in Section \ref{sec:agent} is compared with three other
  policy gradient based methods. The first alternative method is the same as the Agent method, with the only difference that the loss is defined on an action basis rather than on an episodic one,
  resulting in the cost function:
  \begin{equation*}
    J(\Theta) = [\sum_{t=0}^T{(\log \pi_{Prior}(a_t, s_t) - \log \pi_{\Theta}(a_t, s_t))} + \sigma S(A)]^2
  \end{equation*}
    We refer to this method as `Action basis'. The second alternative is a
  REINFORCE algorithm with a reward of $S(A)$ given at the last step. This method is similar to the one used by Silver \textit{et al.} to train the policy network in AlphaGo \cite{Silver2016}, as well as
  the method used by Yu \textit{et al.} \cite{Yu2016}. We refer to this method as `REINFORCE'. The corresponding cost function can be written as:
  \begin{equation*}
    J(\Theta) = S(A)\sum_{t=0}^T \log \pi_{\Theta}(a_t, s_t)
  \end{equation*}
  A variation of this method that considers prior likelihood is defined by changing the reward from $S(A)$ to $S(A)+
  \log P(A)_{Prior}$. This method is referred to as `REINFORCE + Prior', with the cost function:
  \begin{equation*}
    J(\Theta) = [\log P(A)_{Prior} + \sigma S(A)]\sum_{t=0}^T \log \pi_{\Theta}(a_t, s_t)
  \end{equation*}
 Note that the last method by nature strives to generate only the putative sequence with the highest reward. In contrast to the Agent, the optimal policy for this method is not stochastic. This tendency could be restrained by introducing a regularizing policy entropy term. However, it was found that such regularization undermined the models ability to
  produce valid SMILES. This method is therefor dependent on only training sufficiently long for the model to reach a point where highly scored sequences are generated, without being settled
  in a local minima. The experiment aims to answer the following questions:
  \begin{itemize}
    \item{Can the models achieve the task of generating valid SMILES corresponding to structures that do not contain sulphur?}
    \item{Will the models exploit the reward function by converging on na\"{i}ve solutions such as 'C' if not imposed handwritten rules?}
    \item{Are the distributions of physical chemical properties for the generated structures similar to those of sulphur free structures generated by the Prior?}

    \end{itemize}
  
The task is defined by the following scoring function:
\begin{equation*}
  S(A) = \begin{cases}
    \hphantom{-}1 & \text{if valid and no S} \\
    \hphantom{-}0 & \text{if not valid} \\
    -1 & \text{if contains S}
  \end{cases}
\end{equation*}
All the models were  trained for 1000 steps starting from the Prior and 12800 SMILES sequences were
sampled from all the models as well as the Prior. A learning rate of 0.0005 was used for the Agent
  and Action basis methods, and 0.0001 for the two REINFORCE methods. The values of $\sigma$ used were 2 for the Agent and 'REINFORCE + Prior', and 8 for 'Action basis'.
 To explore what effect the training has on the structures
generated, relevant properties for non sulphur containing structures generated by both the Prior and  the other models were compared. The molecular weight,
cLogP, the number of rotatable bonds, and the number of aromatic rings were all calculated using RDKit. The experiment was repeated three times with different random seeds. The
  results are shown in Table \ref{table:sulphur} and randomly selected SMILES generated by the Prior and the different models can be seen in Table \ref{table:sulphursmiles}. For the 'REINFORCE'
method, where the sole aim is to generate valid SMILES that do not contain sulphur, the model quickly learns to exploit the reward funtion by generating sequences containing predominately `C`. This is not surprising,
since any sequence consisting only of this token always gets rewarded. For the 'REINFORCE + Prior'
method, the inclusion of the prior likelihood in the reward function means that this is no longer a viable strategy (the sequences would be given a low prior probability). The model instead tries
to find the structure with the best combination of score and prior likelihood, but as is evident from the SMILES generated and the statistics shown in Table \ref{table:sulphur}, this results in
small, simplistic structures being generated. Thus, both REINFORCE algorithms managed to achieve high scores according to the scoring function, but poorly represented the Prior. Both the Agent and the
'Action basis' methods have explicitly specified target policies. For the 'Action basis' method the policy is specified exactly on a stepwise level,
while for the Agent the target policy is only specified to the likelihoods of entire sequences. Although the 'Action basis' method generates structures that are more similar to the Prior
than the two REINFORCE methods, it performed worse than the Agent despite the higher value of $\sigma$ used while also being slower to learn. This may be due to the less restricted target policy of
the Agent, which could facilitate optimization. The Agent achieved the same fraction of sulphur free structures as the REINFORCE algorithms, while seemingly doing a much better job of representing the Prior. This is
indicated by the similarity of the properties of the generated structures shown in Table \ref{table:sulphur} as well as the SMILES themselves shown in Table \ref{table:sulphursmiles}.

\subsection{Similarity guided structure generation}

\begin{figure}
\includegraphics{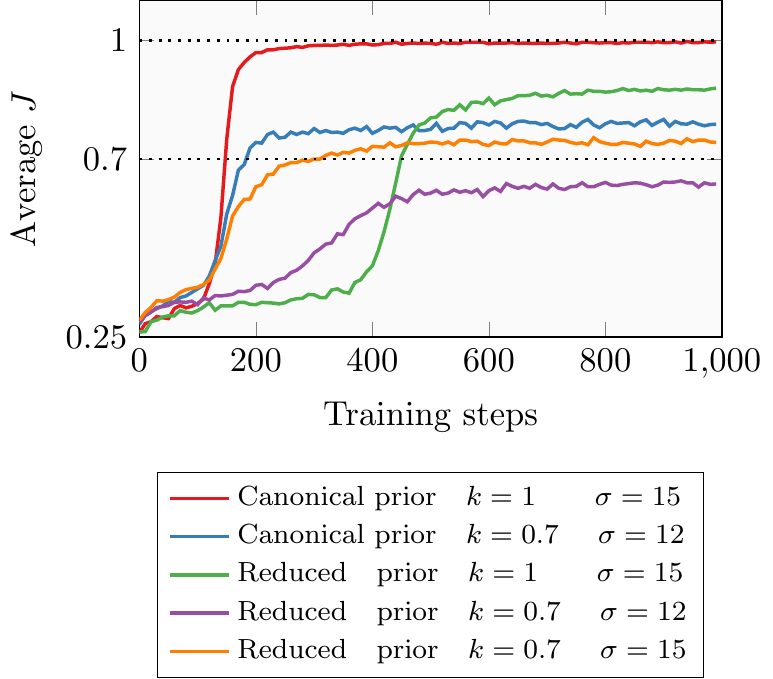}
\caption{\csentence{Average similarity $J$ of generated structures as a function of training steps.} Difference in learning dynamics for the Agents based on the canonical Prior, and those based on a reduced Prior where everything more similar than $J=0.5$ to Celecoxib have been removed.}
\label{fig:tanimoto}
\end{figure}

The second task investigated was that of generating structures similar to a query structure. The Jaccard index \cite{Jaccard1901} $J_{i, j}$ of the RDKit implementation of FCFP4 \cite{Rogers2010}
fingerprints was used as a similarity measure between molecules $i$ and $j$. Compared to the DRD2 activity model (Section \ref{sec:DRD2}), the feature invariant version of the fingerprints and the
smaller diameter 4 was used in order to get a more fuzzy similarity measure. The scoring function was defined as:
\begin{equation*}
  S(A) = -1 + 2 \times \frac{\min \{ J_{i, j}, k \}}{k}
\end{equation*}
This definition means that an increase in similarity is only rewarded up to the point of $k\in[0, 1]$, as well as scaling the reward from $-1$ (no overlap in the fingerprints between query and
generated structure) to $1$ (at least $k$ degree of overlap). Celecoxib was chosen as our query structure, and we first investigated whether Celecoxib itself could be generated by using the high
values of $k=1$ and $\sigma=15$. The Agent was trained for 1000 steps. After a 100 training steps the Agent starts to generate Celecoxib, and after 200 steps it predominately generates this structure
(Figure \ref{fig:tanimoto}).

\begin{figure*}
\includegraphics{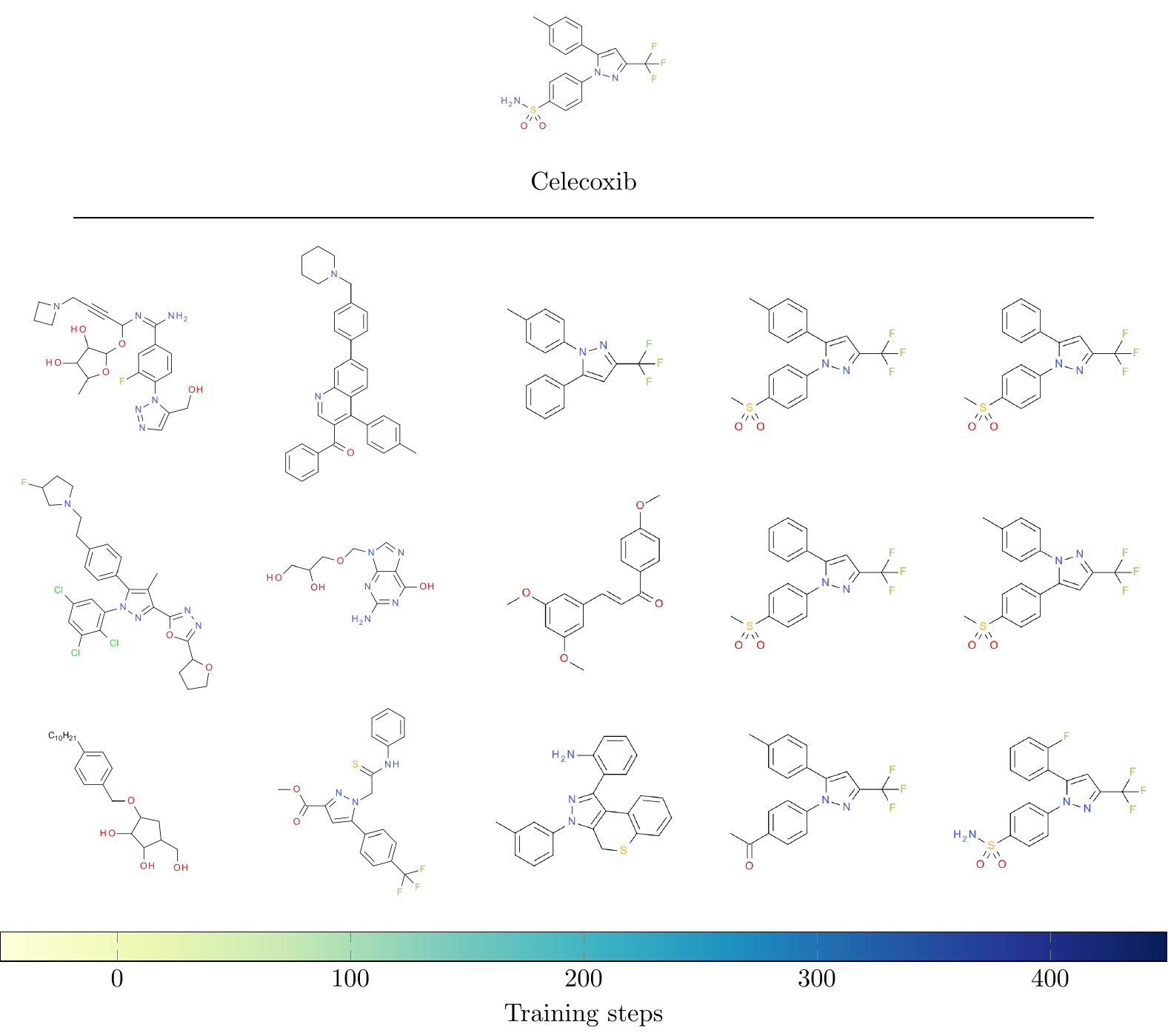}
\caption{\csentence{Evolution of generated structures during training} Structures sampled every 100 training steps during the training of the Agent towards similarity to Celecoxib with $k=0.7$ and $\sigma=15$.}
\label{fig:celecoxib}
\end{figure*}

Celecoxib itself as well as many other similar structures appear in the ChEMBL training set used to build the Prior. An interesting question is whether the Agent could succeed in generating Celecoxib
when these structures are not part of the chemical space covered by the Prior. To investigate this, all structures with a similarity to Celecoxib higher than 0.5 (corresponding to 1804 molecules) were
removed from the training set and a new reduced Prior was trained. The prior likelihood of Celecoxib for the canonical and reduced Priors was compared, as well as the ability of the models to generate
structures similar to Celecoxib. As expected, the prior probability of Celecoxib decreased when similar compounds were removed from the training set from $\log_e P = -12.7$ to $\log_e P = -19.2$,
representing a reduction in likelihood of a factor of 700. An Agent was then trained using the same hyperparameters as before, but on the reduced Prior. After 400 steps, the Agent again managed to
find Celecoxib, albeit requiring more time to do so. After 1000 steps, Celecoxib was the most commonly generated structure (about a third of the generated structures), followed by demethylated
Celecoxib (also a third) whose SMILES is more likely according to the Prior with $\log_e P = -15.2$ but has a lower similarity ($J = 0.87$), resulting in an augmented likelihood equal to that of
Celecoxib.

These experiments demonstrate that the Agent can be optimized using fingerprint based Jaccard similarity as the objective, but making copies of the query structure is hardly useful. A more useful
example is that of generating structures that are moderately to the query structure. The Agent was therefore
trained for 3000 steps, starting from both the canonical as well as the reduced Prior, using $k = 0.7$ and $\sigma = 12$. The Agents based on the canonical Prior quickly converge to their targets,
while the Agents based on the reduced Prior converged more slowly. For the Agent based on the reduced Prior where $k=1$, the fact that Celecoxib and demethylated Celecoxib are given similar augmented
likelihoods means that the average similarity converges to around 0.9 rather than 1.0. For the Agent based on the reduced Prior where $k=0.7$, the lower prior likelihood of compounds similar to
Celecoxib translates to a lower augmented likelihood, which lowers the average similarity of the structures generated by the Agent.

To explore whether this reduced prior likelihood could be offset with a higher value of $\sigma$, an Agent starting from the reduced Prior was trained using $\sigma=15$. Though taking slightly more
time to converge than the Agent based on the canonical Prior, this Agent too could converge to the target similarity. The learning curves for the different model is shown in Figure \ref{fig:tanimoto}.

An illustration of how the type of structures generated by the Agent evolves during training is shown in Figure \ref{fig:celecoxib}. For the Agent based on the reduced Prior with $k=0.7$ and
$\sigma=15$, three structures were randomly sampled every 100 training steps from step 0 up to step 400. At first, the structures are not similar to Celecoxib. After 200 steps, some features from
Celecoxib have started to emerge, and after 300 steps the model generates mostly close analogues of Celecoxib.

We have investigated how various factors affect the learning behaviour of the Agent. In real drug discovery applications, we might be more interested in finding structures with modest similarity to
our query molecules rather than very close analogues. For example, one of the structures sampled after 200 steps shown in Figure \ref{fig:celecoxib} displays a type of scaffold hopping where the
sulphur functional group on one of the outer aromatic rings has been fused to the central pyrazole. The similarity to Celecoxib of this structure is $0.4$, which may be a more interesting solution for
scaffold-hopping purposes. One can choose hyperparameters and similarity criterion tailored to the desired output. Other types of similarity measures such as pharmacophoric fingerprints
\cite{Reutlinger2013}, Tversky substructure similarity \cite{Senger2009}, or presence/absence of certain pharmacophores could also be explored.

\subsection{Target activity guided structure generation}
\label{subsec:drd2}

The third example, perhaps the one most interesting and relevant for drug discovery, is to optimize the Agent towards generating structures with predicted biological activity. This can be seen as a
form of inverse QSAR, where the Agent is used to implicitly map high predicted probability of activity to molecular structure. DRD2 was chosen as the biological target. The clustering split of the
DRD2 activity dataset as described in Section \ref{sec:DRD2} resulted in 1405, 1287, and 4526 actives in the test, validation, and training sets respectively. The average 
similarity to the nearest neighbour in the training set for the test set compounds was 0.53. For a random split of actives in sets of the same sizes this 
similarity was 0.69, indicating that the clustering had significantly decreased training-test set similarity which mimics the hit finding practice in drug discovery to identify
diverse hits to the training set. Most of the DRD2 actives are also included in the ChEMBL dataset used to train the Prior. To explore the effect of not having the known actives included in the Prior,
a reduced Prior was trained on a reduced subset of the ChEMBL training set where all DRD2 actives had been removed.

\begin{table}
  \centering
  \caption{Performance of the DRD2 activity model}
  \begin{tabular}{lccc}
    \toprule
    Set & Training & Validation & Test\\
    \midrule
    Accuracy & 1.00 & 0.98 & 0.98\\
    ROC-AUC & 1.00 & 0.99 & 1.00\\
    Precision & 1.00 & 0.96 & 0.97\\
    Recall & 1.00 & 0.73 & 0.82\\
    \bottomrule
  \end{tabular}
  \label{table:svm_table}
\end{table}

\begin{table*}
  \centering
  \caption{Comparison of properties for structures generated by the canonical Prior, the reduced Prior, and corresponding Agents.}
  \begin{tabular}{lcccc}
    \toprule
    Model & Prior & Agent & $\mathrm{Prior}^\dagger$ & $\mathrm{Agent}^\dagger$\\
    \midrule
    Fraction valid SMILES & 0.94 & 0.99 & 0.94 & 0.99\\
    Fraction predicted actives & 0.03 & 0.97 & 0.02 & 0.96\\
    Fraction similar to train active\ \ \ \ \  & 0.02 & 0.79 & 0.02 & 0.75\\
    Fraction similar to test active & 0.01 & 0.46 & 0.01 & 0.38\\
    Fraction of test actives recovered ($\times 10^{-3}$) & 13.5 & 126 & 2.85 & 72.6\\
    Probability of generating a test set active ($\times 10^{-3}$)\ \ \ \ \ \ & 0.17 & 40.2 & 0.05 & 15.0\\
    \bottomrule
  \end{tabular}
  \smallskip \linebreak $^{\dagger}$DRD2 actives witheld from the training of the Prior
  \label{table:DRD2}
\end{table*}

\begin{figure*}
\includegraphics{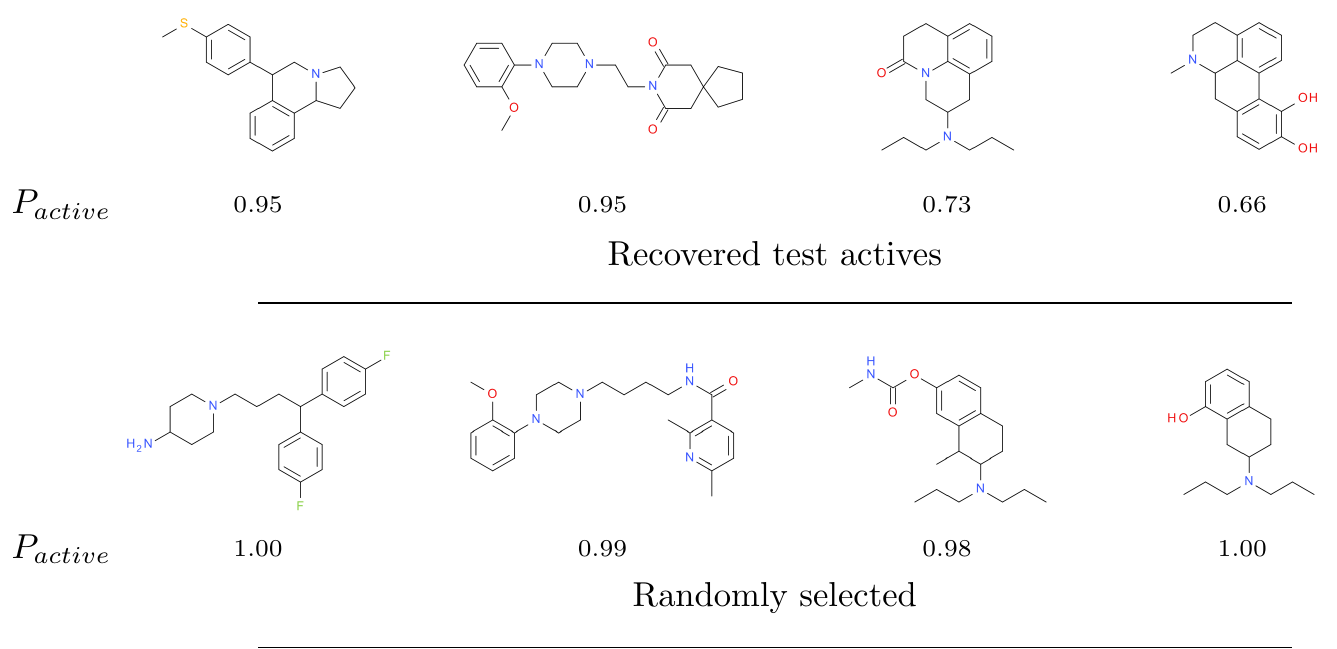}
\caption{\csentence{Structures designed by the Agent to target DRD2.} Molecules generated by the Agent based on the reduced Prior. On the top are four of the test set actives that were recovered, and below are four randomly selected structures. The structures are annotated with the predicted probability of being active.}
\label{fig:DRD2_gen}
\end{figure*}

The optimal hyperparameters found for the SVM activity model were $C=2^{7}, \gamma=2^{-6}$, resulting in a model whose performance is shown in Table \ref{table:svm_table}. The good performance in
general can be explained by the apparent difference between actives and inactive compounds as seen during the clustering, and the better performance on the test set compared to the validation set
could be due to slightly higher nearest neighbour similarity to the training actives (0.53 for test actives and 0.48 for validation actives).

The output of the DRD2 model for a given structure is an uncalibrated predicted probability of being active $P_{active}$. This value is used to formulate the following scoring function:
\begin{equation*}
  S(A) = -1 + 2 \times P_{active}
\end{equation*}
The model was trained for 3000 steps using $\sigma = 7$. After training, the fraction of predicted actives according to the DRD2 model increased from 0.02 for structures generated by the reduced Prior
to 0.96 for structures generated by the corresponding Agent network (Table \ref{table:DRD2}). To see how well the structure-activity-relationship learnt by the activity model is transferred to the
type of structures generated by the Agent RNN, the fraction of compounds with an ECFP6 Jaccard similarity greater than 0.4 to any active in the training and test sets was calculated.

\begin{figure*}
\includegraphics{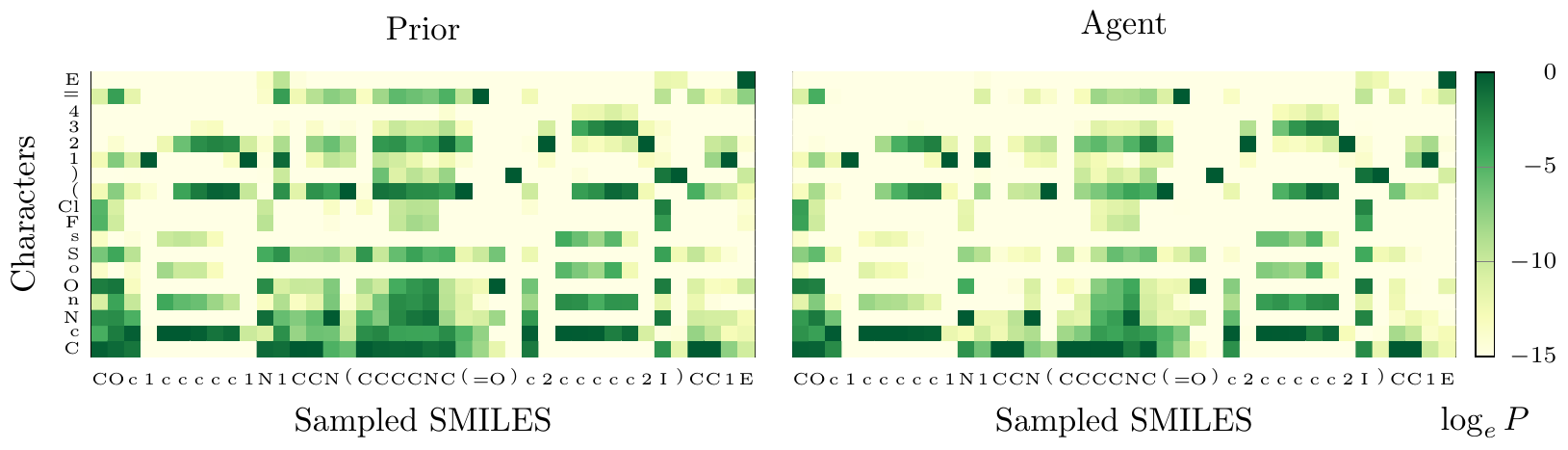}
\caption{\csentence{A (small) change of mind.} The conditional probability distributions when the DRD2 test set active 'COc1ccccc1N1CCN(CCCCNC(=O)c2ccccc2I)CC1' is generated by the Prior and an Agent trained using the DRD2 activity model, for the case where all actives used to build the activity model have been removed from the Prior. E = EOS.}
\label{fig:heatmap2}
\end{figure*}

In some cases, the model recovered exact matches from the training and test sets (c.f. Segler \textit{et al.} \cite{Segler2017}). The fraction of recovered test actives recovered by
the canonical and reduced Prior were 1.3\% and 0.3\% respectively. The Agent derived from the canonical Prior managed to recover 13\% test
actives; the Agent derived from the reduced Prior recovered 7\%. For the Agent derived from the reduced Prior, where the DRD2 actives were excluded from the Prior training set, this means that the
model has learnt to generate "novel" structures that have been seen neither by the DRD2 activity model nor the Prior, and are experimentally confirmed actives. We can formalize this observation by
calculating the probability of a given generated sequence belonging to the set of test actives. For the canonical and reduced Priors, this probability was 0.17$\times 10^{-3}$ and 0.05$\times 10^{-3}$
respectively. Removing the actives from the Prior thus resulted in a threefold reduction in the probability of generating a structure from the set of test actives. For the Agents, the probabilities
rose to 15.0$\times 10^{-3}$ and 40.2$\times 10^{-3}$ respectively, corresponding to an enrichment of a factor of 250 over the Prior models. Again the consequence of removing the actives from the
Prior was a threefold reduction in the probability of generating a test set active: the difference between the two Priors is directly mirrored by their corresponding Agents. Apart
  from generating a higher
fraction of structures that are predicted to be active, both Agents also generate a significantly higher fraction of valid SMILES (Table \ref{table:DRD2}). Sequences that are not valid SMILES receive a score of -1, which
means that the scoring function naturally encourages valid SMILES. 

A few of the test set actives generated by the Agent based on the reduced Prior along with a few randomly selected generated structures are shown together with their predicted probability of activity
in Figure \ref{fig:DRD2_gen}. Encouragingly, the recovered test set actives vary considerably in their structure, which would not have been the case had the Agent converged to generating only a
certain type of very similar predicted active compounds.

Removing the known actives from the training set of the Prior resulted in an Agent which shows a decrease in all metrics measuring the overlap between the known actives and the structures generated,
compared to the Agent derived from the canonical Prior. Interestingly, the fraction of predicted actives did not change significantly. This indicates that the Agent derived from the reduced Prior has
managed to find a similar chemical space to that of the canonical Agent, with structures that are equally likely to be predicted as active, but are less similar to the known actives. Whether or not
these compounds are active will be dependent on the accuracy of the target activity model. Ideally, any predictive model to be used in conjunction with the generative model should cover a broad
chemical space within its domain of applicability, since it initially has to assess representative structures of the dataset used to build the Prior \cite{Segler2017}.

Figure \ref{fig:heatmap2} shows a comparison of the conditional probability distributions for the reduced Prior and its corresponding Agent when a molecule from the set of test actives is generated.
It can be seen that the changes are not drastic with most of the trends learnt by the Prior being carried over to the Agent. However, a big change in the probability distribution even only at one step
can have a large impact on the likelihood of the sequence and could significantly alter the type of structures generated.

\section{Conclusion}
To summarize, we believe that an RNN operating on the SMILES representation of molecules is a promising method for molecular \textit{de novo} design. It is a data-driven generative model that does not
rely on pre-defined building blocks and rules, which makes it clearly differentiated from traditional methods. In this study we extend upon previous work \cite{Segler2017, Bombarelli2016, Yu2016,
  Jaques2016} by introducing a reinforcement learning method which can be used to tune the RNN to generate structures with certain desirable properties through augmented episodic
likelihood.

The model was tested on the task of generating sulphur free molecules as a proof of principle, and the method using augmented episodic likelihood was compared with traditional policy gradient methods. 
The results indicate that the Agent can find solutions reflecting the
underlying probability distribution of the Prior, representing a significant improvement over both traditional REINFORCE \cite{Williams1992} algorithms as well as previously reported methods \cite{Jaques2016}. To evaluate if the model could be used to generate analogues to a query structure, the Agent was trained to generate structures similar to the
drug Celecoxib. Even when all analogues of Celecoxib were removed from the Prior, the Agent could still locate the intended region of chemical space which was not part of the Prior. Further more, when
trained towards generating predicted actives against the dopamine receptor type 2 (DRD2), the Agent generates structures of which more than 95\% are predicted to be active, and could recover test set actives
even in the case where they were not included in either the activity model nor the Prior. Our results indicate that the method could be a useful tool for drug discovery.

It is clear that the qualities of the Prior are reflected in the corresponding Agents it produces. An exhaustive study which explores how parameters such as training set size, model size,
regularization \cite{Zaremba2014, Wan2013}, and training time would influence the quality and variety of structures generated by the Prior would be interesting. Another interesting avenue for future
research might be that of token embeddings \cite{Bengio2003}. The method was found to be robust with respect to the hyperparameters $\sigma$ and the learning rate.

All of the aforementioned examples used single parameter based scoring functions. In a typical drug discovery project, multiple parameters such as target activity, DMPK profile, synthetic
accessibility etc. all need to be taken into account simultaneously. Applying this type of multi-parametric scoring functions to the model is an area requiring further research.

\begin{backmatter}

  \section*{Additional Files}
  \subsection*{Additional file 1 --- Equivalence to REINFORCE}
  Proof that the method used can be described as a REINFORCE type algorithm.
  \subsection*{Additional file 2 --- Generated structures}
  Structures generated by the canonical Prior and different Agents.

\section*{Availability of data and materials}
The source code and data supporting the conclusions of this article is available at https://github.com/MarcusOlivecrona/REINVENT, DOI:10.5281/zenodo.572576.

\begin{itemize}
\item{Project name: REINVENT}
\item{Project home page: https://github.com/MarcusOlivecrona/REINVENT}
\item{Archived version: http://doi.org/10.5281/zenodo.572576}
\item{Operating system: Platform independent}
\item{Programming language: Python}
\item{Other requirements: Python2.7, Tensorflow, RDKit, Scikit-learn}
\item{License: MIT}
\end{itemize}

\section*{Declarations}

\subsection*{Ethics approval and consent to participate}
Not applicable.
  
\subsection*{Consent for publication}
Not applicable.
  
\subsection*{List of abbreviations}
\begin{itemize}
\item{DMPK - Drug metabolism and pharmacokinetics}
\item{DRD2 - Dopamine receptor D2}
\item{QSAR - Quantitive structure activity relationship}
\item{RNN - Recurrent neural network}
\item{RL - Reinforcement Learning}
\item{Log - Natural logarithm}
\item{BPTT - Back-propagation through time}
\item{$A$ - Sequence of tokens constituting a SMILES}
\item{Prior - An RNN trained on SMILES from ChEMBL used as a starting point for the Agent}
\item{Agent - An RNN derived from a Prior, trained using reinforcement learning}
\item{$J$ - Jaccard index}
\item{ECFP6 - Extended Molecular Fingerprints with diameter 6}
\item{SVM - Support Vector Machine}
\item{FCFP4 - Extended Molecular Fingerprints with diameter 4 and feature invariants}
\end{itemize}

\subsection*{Competing interests}
The authors declare that they have no competing interests.

\subsection*{Funding}
MO, HC, and OE are employed by AstraZeneca. TB has received funding from the European Union's Horizon 2020 research and innovation program under the Marie Sklodowska-Curie grant agreement No 676434,
"Big Data in Chemistry" ("BIGCHEM", http://bigchem.eu). The article reflects only the authors' view and neither the European Commission nor the Research Executive Agency (REA) are responsible for
any use that may be made of the information it contains.

\subsection*{Author's contributions}
MO contributed concept and implementation. All authors co-designed experiments. All authors contributed to the interpretation of results. MO wrote the manuscript. HC, TB, and OE reviewed and edited
the manuscript. All authors read and approved the final manuscript.

\subsection*{Acknowledgements}
The authors thank Thierry Kogej and Christian Tyrchan for general assistance and discussion, and Dominik Peters for his expertise in \LaTeX.

\bibliographystyle{bmc-mathphys}
\bibliography{bibliography.bib}

\end{backmatter}

\clearpage
\includepdf[pages={1}]{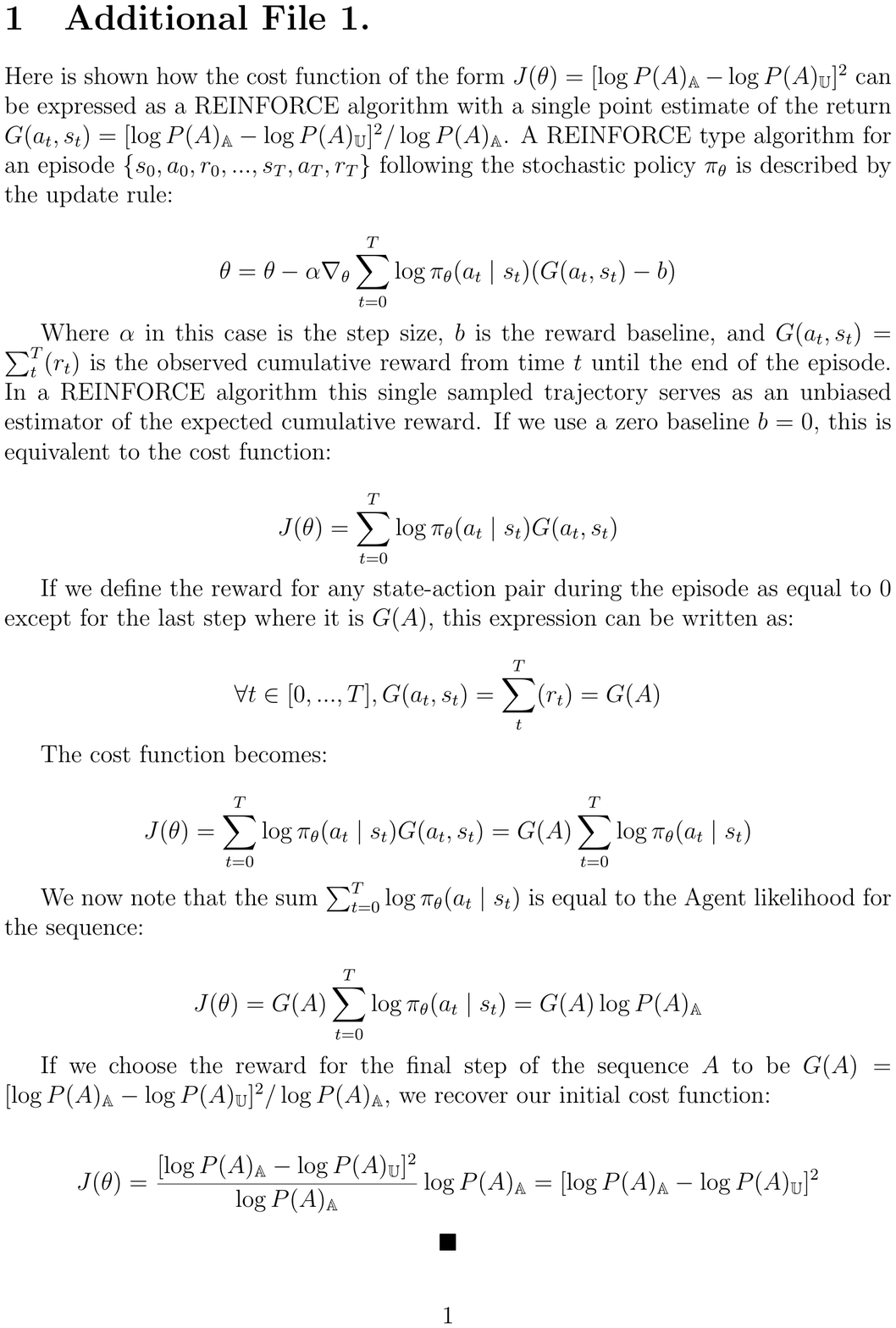}

\begin{figure*}[!hb]
\includegraphics[scale=0.7]{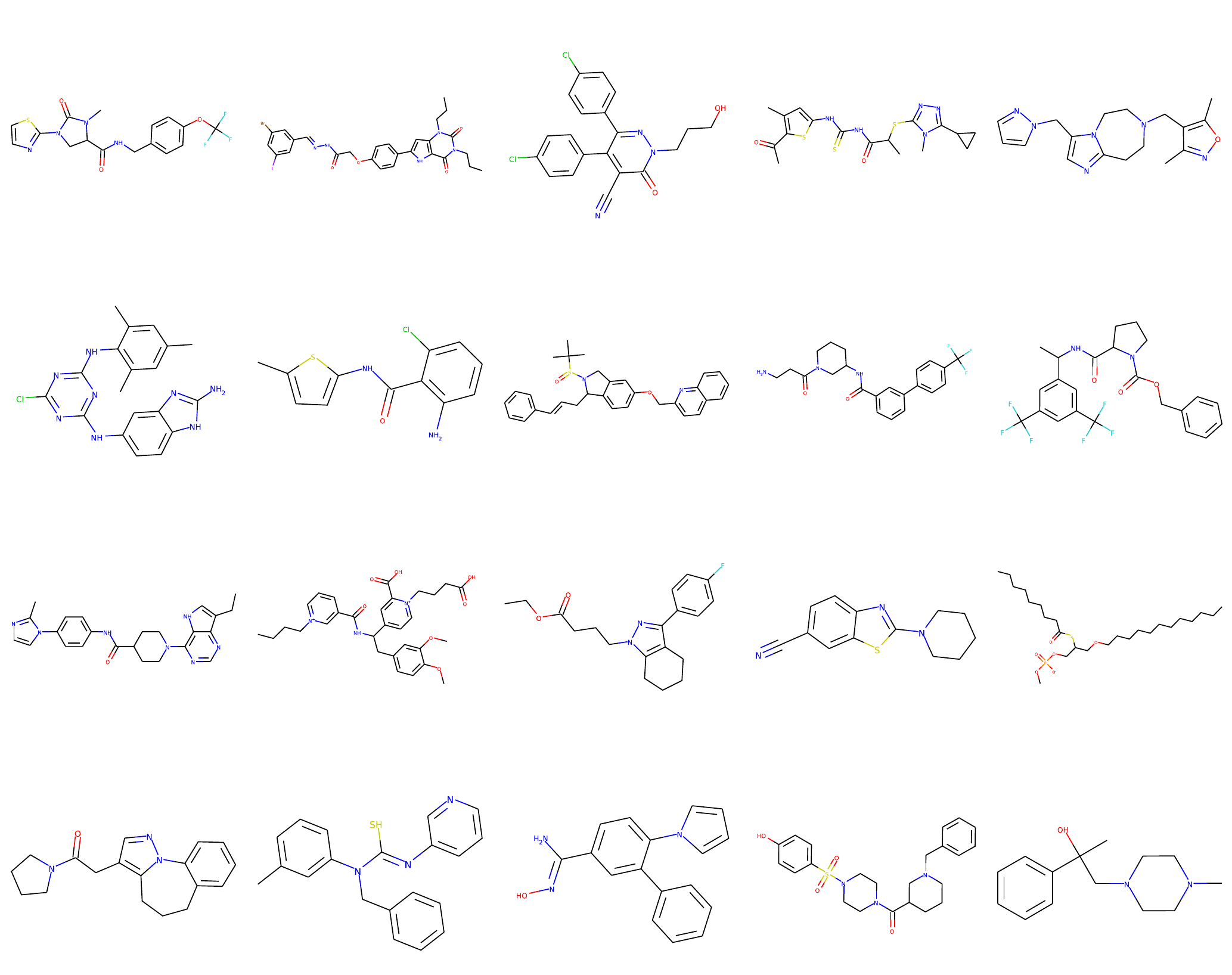}
\caption{\csentence{Additional file 2.1} Randomly selected structures generated by the canonical Prior.}
\end{figure*}

\begin{figure*}
\includegraphics[scale=0.7]{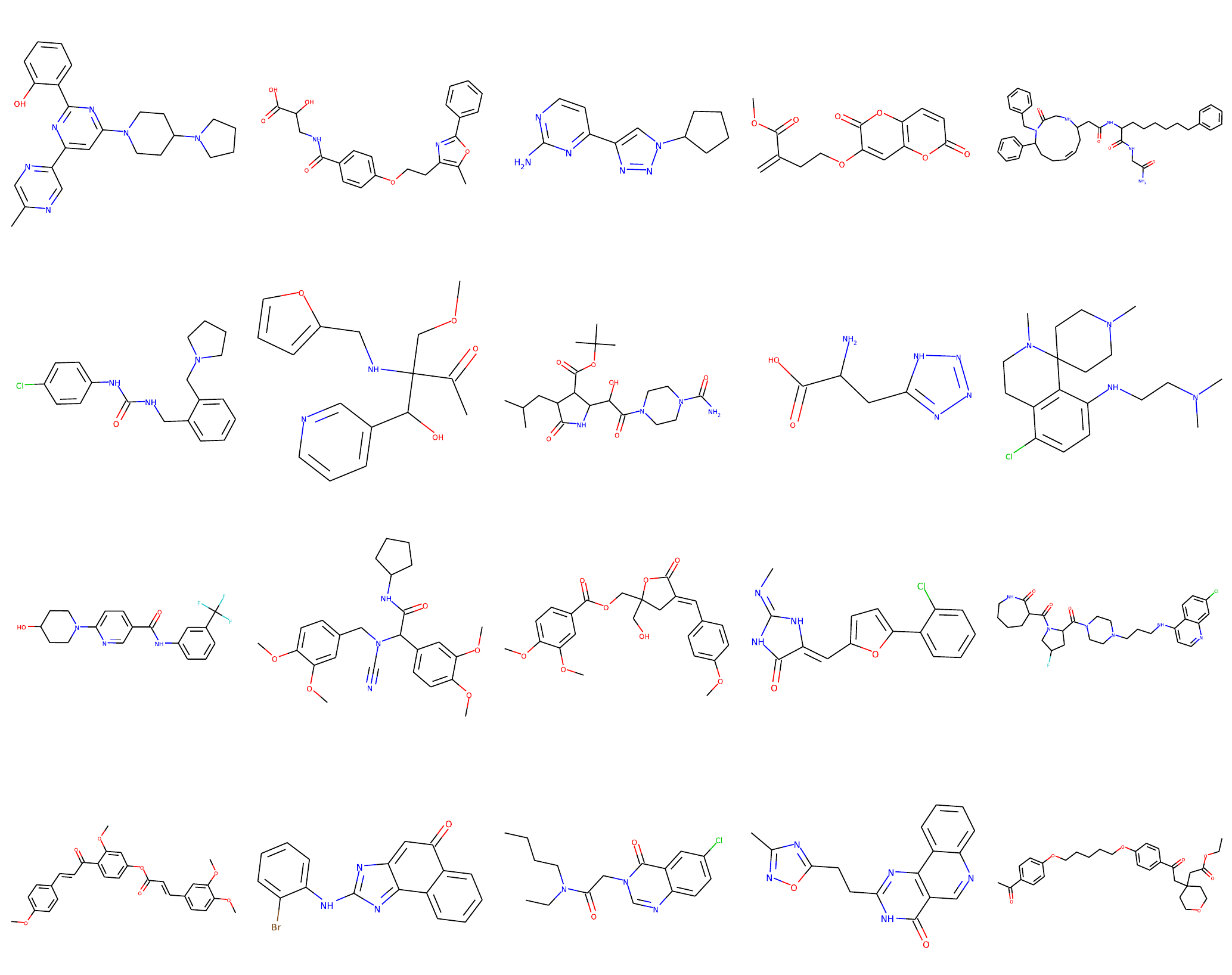}
\caption{\csentence{Additional file 2.2} Randomly selected structures generated by the Agent trained to avoid sulphur.}
\end{figure*}

\begin{figure*}
\includegraphics[scale=0.7]{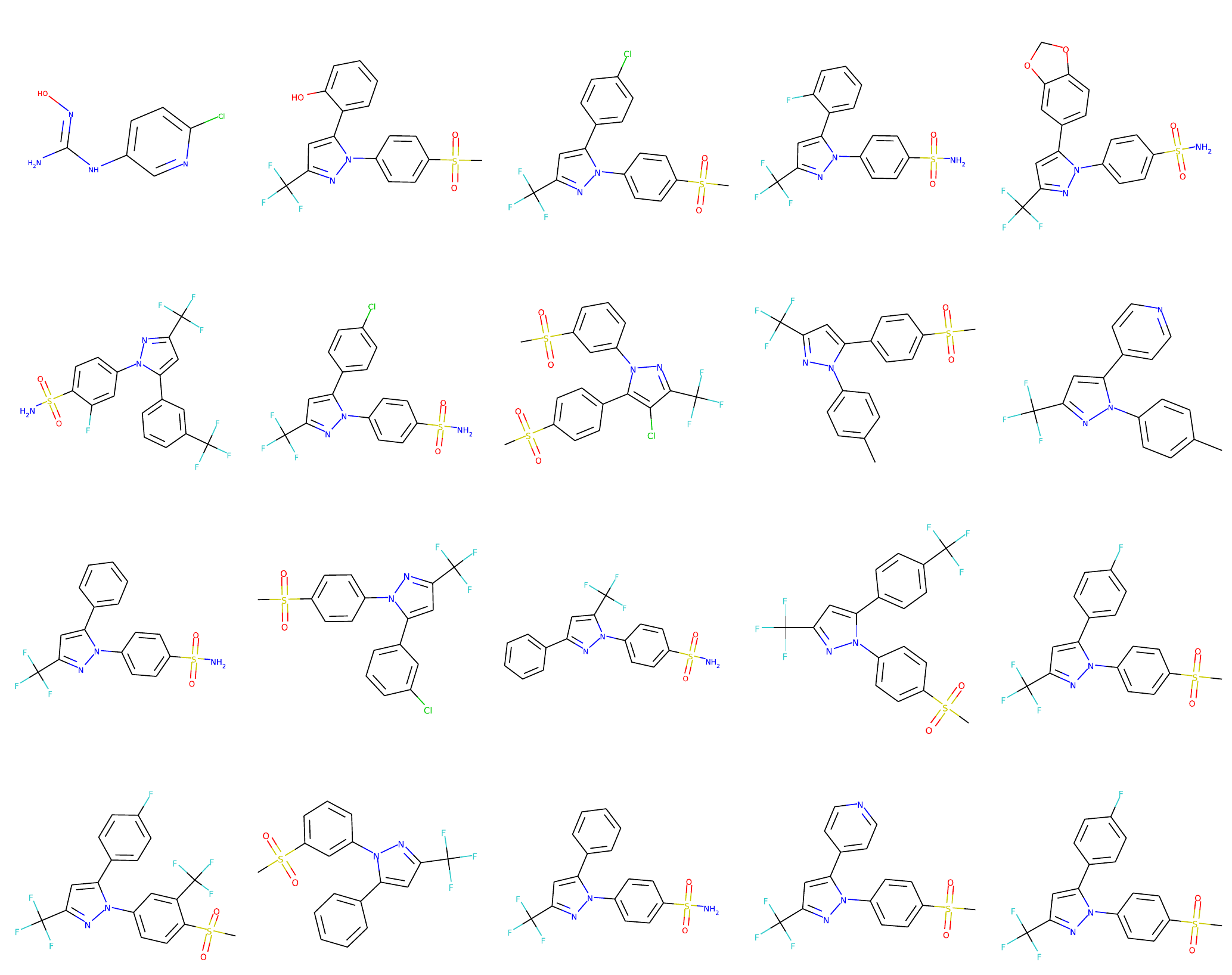}
\caption{\csentence{Additional file 2.3} Randomly selected structures generated by the Agent based on the reduced Prior trained to design analogues of Celecoxib.}
\end{figure*}

\begin{figure*}
\includegraphics[scale=0.7]{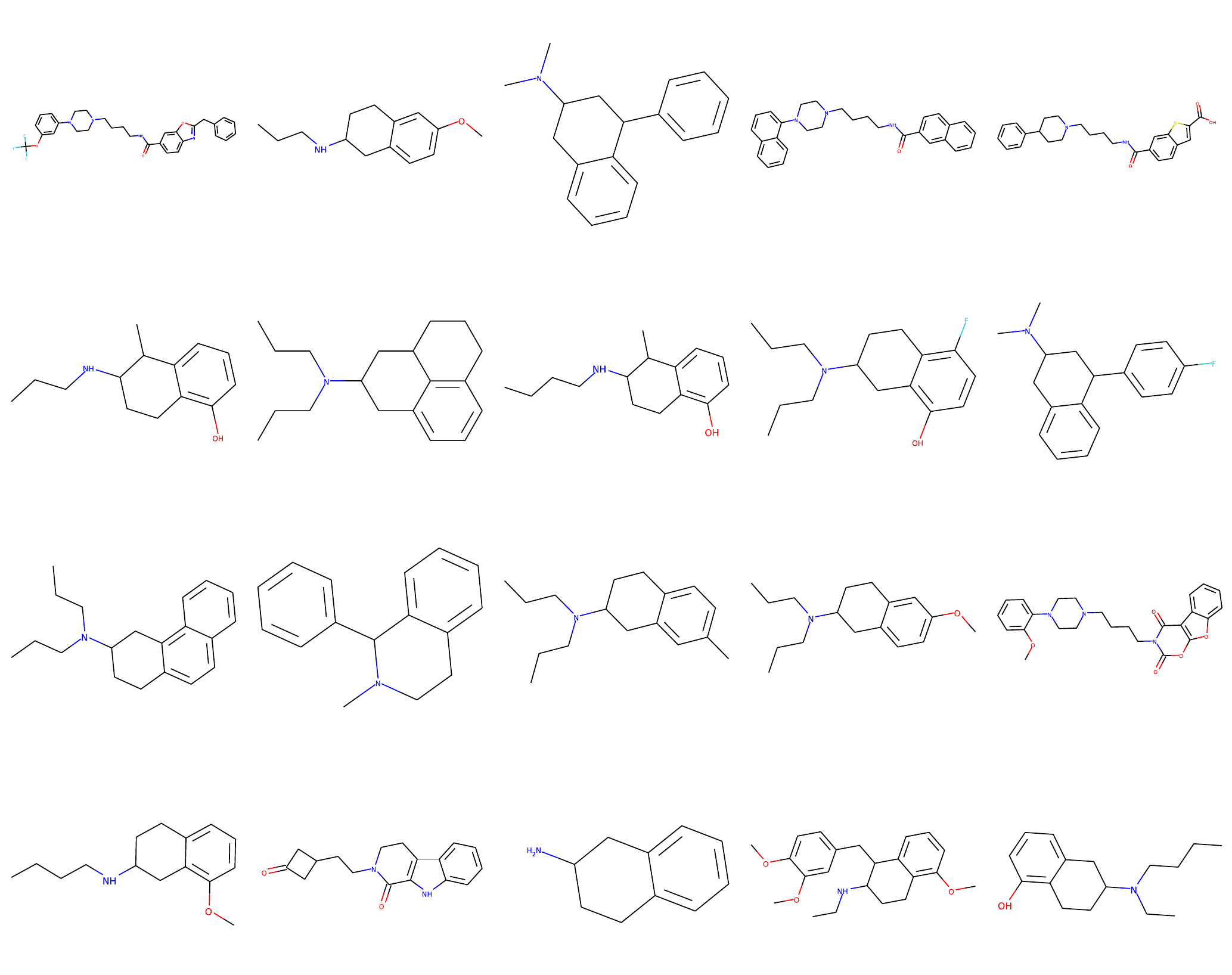}
\caption{\csentence{Additional file 2.4} Randomly selected structures generated by the Agent based on the reduced Prior trained to design actives against DRD2.}
\end{figure*}
\end{document}